\documentclass[letterpaper, 10 pt, conference]{ieeeconf}  % Comment this line out if you need a4paper

\IEEEoverridecommandlockouts                              % This command is only needed if 
\usepackage{graphicx} % for pdf, bitmapped graphics files

\title{\LARGE \bf
UAV Visual Teach and Repeat Using Only Semantic Object Features
}

\author{Amirmasoud Ghasemi Toudeshki, Faraz Shamshirdar and Richard Vaughan$^{1}$% <-this % stops a space
\thanks{$^{1}$Autonomy Lab, School of Computing Science, Simon Fraser University, Burnaby, BC, Canada
        {\tt\small \{ghasemit,fshamshi,vaughan\}@sfu.ca}}%
}

\begin{document}

\maketitle
\thispagestyle{empty}
\pagestyle{empty}

%%%%%%%%%%%%%%%%%%%%%%%%%%%%%%%%%%%%%%%%%%%%%%%%%%%%%%%%%%%%%%%%%%%%%%%%%%%%%%%%
\begin{abstract}
We demonstrate the use of semantic object detections as robust features for Visual Teach and Repeat (VTR). Recent CNN-based object detectors are able to reliably detect objects of tens or hundreds of categories in video at frame rates. We show that such detections are repeatable enough to use as landmarks for VTR, without any low-level image features. Since object detections are highly invariant to lighting and surface appearance changes, our VTR can cope with global lighting changes and local movements of the landmark objects. In the teaching phase we build a series of compact scene descriptors: a list of detected object labels and their image-plane locations. In the repeating phase, we use Seq-SLAM-like relocalization to identify the most similar learned scene, then use a motion control algorithm based on the funnel lane theory to navigate the robot along the previously piloted trajectory.

We evaluate the method on a commodity UAV, examining the robustness of the algorithm to new viewpoints, lighting conditions, and movements of landmark objects. The results suggest that semantic object features could be useful due to their invariance to superficial appearance changes compared to low-level image features.
\end{abstract}

%%%%%%%%%%%%%%%%%%%%%%%%%%%%%%%%%%%%%%%%%%%%%%%%%%%%%%%%%%%%%%%%%%%%%%%%%%%%%%%%
\section{INTRODUCTION}
%why VTR? 
Visual Teach and Repeat (VTR) has become a canonical task in robotics, and has many practical applications including surveillance patrols and transporting goods. The challenge is to conveniently teach and reliably repeat a path in a way that is robust to normal environmental changes such as lighting, weather, and appearance changes due to local activity. An interesting extreme version of the task is to use just a monocular camera: a cheap and ubiquitous sensor that does not rely on ambient infrastructure like GPS. VTR has been studied using land, air and water vehicles. Here we consider monocular VTR on Unmanned Air Vehicles (UAVs). 

%phases 
VTR has two phases. At the teaching phase, a map-like memory is recorded from observations made while the robot (or other training device) is piloted over the desired path. Then at the repeat phase the robot must relocalize itself on the prior path, and repeat the remainder of it based on the observations in the reference memory. The design of the memory data structure is important. The two main approaches are (i) to make a 3D map using SLAM, with landmarks in 3D space; or (ii) to record a sequence of keyframes, with landmarks in image space. We will use object detections in image space to create compact keyframe descriptors.

%why not point features?
Robust VTR requires invariance to small changes in the environment. Here we examine the use of semantic object detections as high-level image descriptors that are invariant to superficial appearance changes. Low-level features have some important drawbacks: direct descriptors which use image patches are sensitive to camera viewpoint and illumination changes. Point features like SIFT \cite{lowe1999object}, SURF \cite{bay2006surf}, ORB \cite{rublee2011orb} and BRIEF \cite{calonder2010brief} are less sensitive to illumination, but are still sensitive to the viewpoint of the camera, particularly in scale. 

%why object convolutional deep network? %why semantic?
Recent deep convolutional neural networks (CNNs) trained on millions of examples provide excellent results in detecting a wide variety of objects under enormous variations in illumination, occlusion and viewpoint. Here we exploit this to find objects in keyframes and use them as landmarks that are re-detected under very large scale, illumination and surface appearance changes, including translating or completely rotating the object, or replacing it with another object of the same category. Our intuition is that discrete, recognizable objects such as `lamppost' and `park bench' should work as robustly salient features for remembering directions, as they do for humans. 

%Semantic description of an environment faciliates the human-robot dialog ***for assigning the tasks to robots*** [Simultaneous Localization and Mapping with Learned Object Recognition and Semantic Data Association]. For example in VTR, one can ask the robot to find and take a picture of a specific object during the repeating phase. Objects also may move, rotate or being substituted with the same class of object but very different in shape or color. As an example, a parked car in a parking lot might substituted by another car. In this case, object level perception could still recognize the car and match the scene in the memory. Because of these benefits, In this article, we propose the concept of semantic VTR. We show that this method has a reliable perfermonce by testing the algorithm on a commercial uav. We run the robot in different initial position and random orientation. We also change the lighting and make some random changes in object poses to show how the performance of the algorithm changes.

\begin{figure}[t!]
	\centering
	\includegraphics[width=0.95\columnwidth]{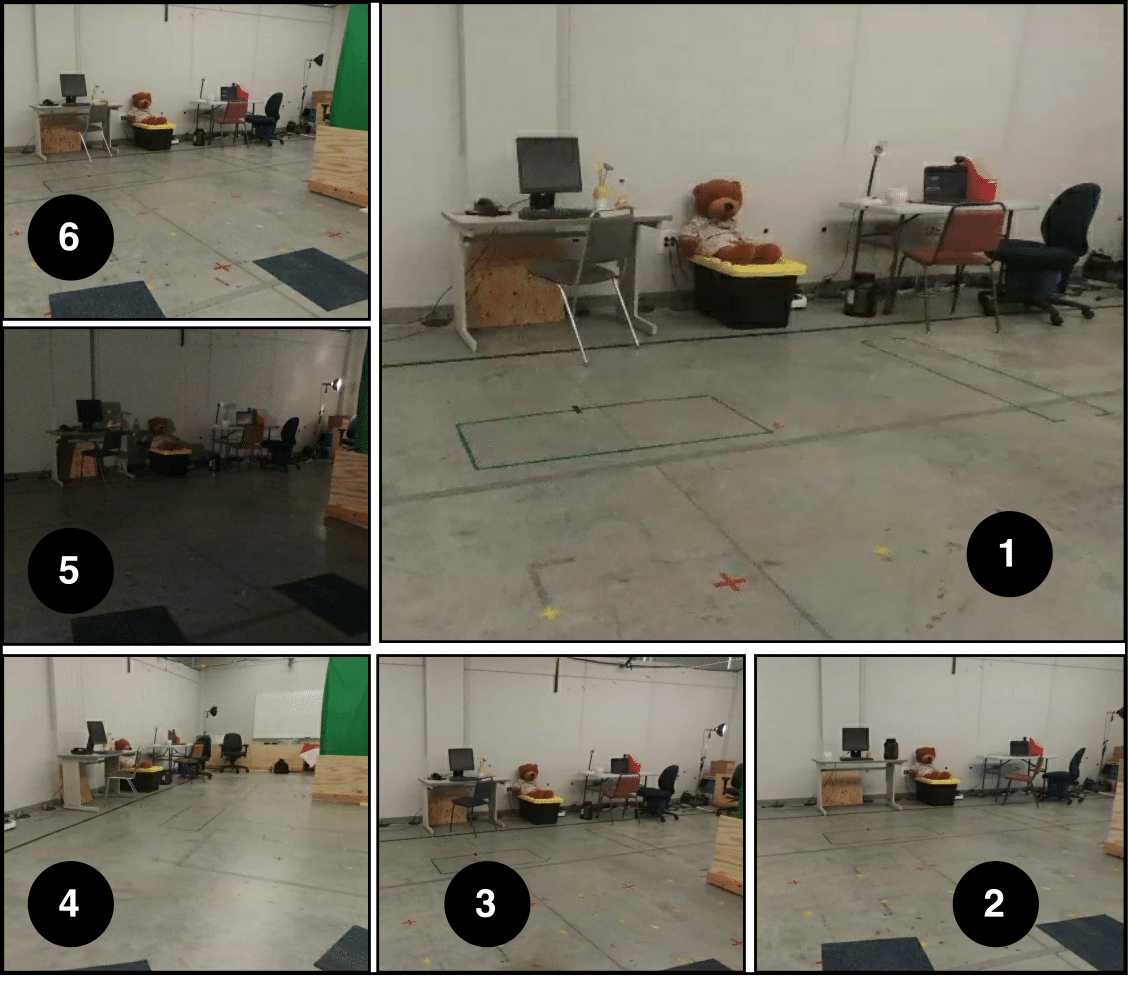}
	\caption{Relocalization robustness: the UAV is able to repeat a learned trajectory despite large changes in viewpoint and scene contents and appearance. (1) is the first image in the taught sequence; (2-6) are examples where the robot has successfully relocalized by matching a sequence of object locations (chair, screen, bear, etc.) from a sequence of descriptors stored during teaching and correctly located (1) as the closest matching image. Note the missing and moved objects in the repeats.} 
	\label{front_fig}
\end{figure}

%Relocalization and control
Our approach to relocalization is similar to appearance-based algorithms but based on high-level features. We use sequential-temporal relocalization inspired by Sequential SLAM \cite{milford2012seqslam}. This reduces the effect of perceptual aliasing caused by a limited landmark vocabulary and number of landmarks by using a sequence of observations instead of one image. The motion controller is based on the funnel control theory \cite{chen2009qualitative} with some extensions to the original implementation.

%Following Sections - RV: this is almost never worthwhile - we all know how papers are structured....

% The proposed methods and conducted tests are brought in details in the following sections. We bring a short overview on the related works in section 2. Then, in section 3, the method for semantic VTR is discussed in detail. Section 4 is describing the experiments that we did for proving the concept of semantic VTR. In section 5, we demonstrate the results and finally in section 6, the conclusion of the work and future ideas are described.

\section{Related work} % work, not works - work is also plural

%UAV-VTR 
%ICRA version
%VTR using UAVs is only recently feasible, with few early examples \cite{pfrunder2014proof} \cite{nguyen2016appearance}. Pfrunder et.al \cite{pfrunder2014proof} demonstrate a simulation-only quad-rotor with downward-facing camera that repeats a simple path starting from the same position with no changes in the environment. The method is similar to \cite{furgale2010visual} and based on SURF descriptors and pose-based localization. Nguyen et al \cite{nguyen2016appearance} also choose SURF features for descriptors but their localization technique is appearance-based. Again, experiments are in simulation, using an AR-Drone model in Gaezbo, and again the environment is unchanged between teach and repeat.

%Modified Version
VTR using UAVs is only recently feasible, with few early examples \cite{pfrunder2014proof} \cite{nguyen2016appearance}. Pfrunder et.al \cite{pfrunder2014proof} demonstrate a quad-rotor with downward-facing camera that repeats a simple path starting from the same position with no changes in the environment. The method is similar to \cite{furgale2010visual} and based on SURF descriptors and pose-based localization. Nguyen et al \cite{nguyen2016appearance} also choose SURF features for descriptors but their localization technique is appearance-based. Experiments are in simulation, using an AR-Drone model in Gaezbo, and again the environment is unchanged between teach and repeat.
%Novel work in UAV-VTR
Surber et al \cite{surber2017robust} build a map using structure from motion (SfM) from data captured in the teaching phase and execute visual-inertial odometry on a real UAV in the repeating phase for localization, using a previously constructed map as prior knowledge. They use BRIEF descriptors and report that relocalization is successful only where the appearance is similar enough to the reference map.

% RV TODO missing ref 'bridging' paper

%Monocular and Barfoot
Barfoot has a significant body of work on Teach and Repeat navigation, including \cite{clement2017robust} which tested a ground robot in a path over 1km, demonstrating a color-constant method to handle changes in illumination. However, this work is sensitive to changes of viewpoint and hence requires the initial position to be similar. To handle gradually-accumulating environmental changes, \cite{mactavish2017visual} employs a `multi-experience' method for VTR, which was introduced in \cite{paton2016bridging}.
%GRIEF
Krajkik et al \cite{krajnik2017image} examines combinations of feature detectors and descriptors for VTR, favouring the STAR feature detector and GRIEF feature descriptor for good accuracy with computational speed. GRIEF is a binary feature descriptor intended to be robust to seasonal changes in appearance. 
%Semantic
So-called `semantic' segmentation of foreground objects from images is a traditional topic in computer vision and robotics  \cite{rogers2011simultaneous}, \cite{salas2013slam++}. The recent development in deep neural networks, particularly CNNs has this area flourishing \cite{sunderhauf2016place,mu2016slam,taisho2016mining,bowman2017probabilistic}. New object detection algorithms provide much better accuracy and speed than was previously available \cite{redmon2016yolo9000,ren2015faster,liu2016ssd} which inspired our attempt to use them as online real-time feature detectors.

\section{Method}
\begin{figure}[t!]
	\centering
	\includegraphics[width=0.95\columnwidth]{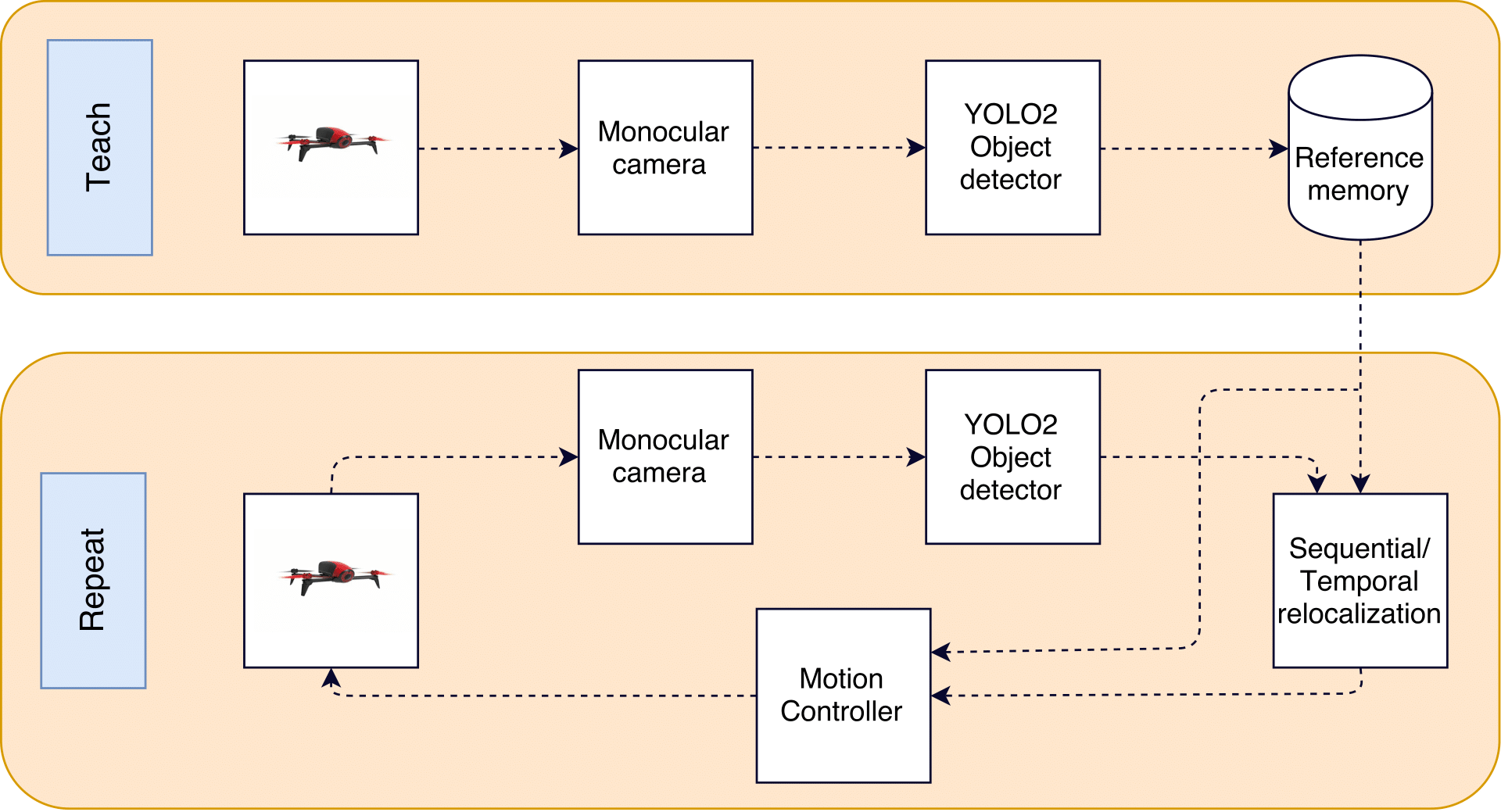}
	\caption{System Overview}
	\label{sys_overview}
\end{figure}
\subsection{System Overview}
% TODO: refactor these facts about teaching and repeating phases.

%General Description
Figure \ref{sys_overview} outlines the structure of our system. During teaching, the UAV is flown or carried along the target trajectory. Video frames from the UAV are periodically input to the YOLO-2 CNN object detector. The CNN output is processed to obtain a scene descriptor of the locations of objects in the image. Each descriptor is stored in sequence. The finished sequence is the robot's memory of the learned trajectory. 

%repeat
In the repeat phase, the UAV motion is actively directly controlled by the VTR system. Again, images from the robot are input to the object detector to obtain a scene descriptor. The relocalization module then finds the closest match between the current and memorized scenes. Once localized, the robot's motion controller attempts to move the robot so that the next scene will look like the next scene in the stored sequence. These modules are described in more detail below.

\subsection{Detection} % RV
%ICRA Vesion:
%We use the YOLO-2 \cite{redmon2016yolo9000} CNN object detector trained on the Pascal VOC dataset. It is notable for its speed in comparison with other deep networks like Faster RCNN \cite{ren2015faster}. We can detect objects at 60fps on a commodity PC with NVIDA-GTX 1080Ti GPU.

%Modified Version:
We use the YOLO-2 \cite{redmon2016yolo9000} CNN object detector trained on the COCO dataset. It is notable for its speed in comparison with other deep networks like Faster RCNN \cite{ren2015faster}. We can detect objects at 40fps on a commodity PC with NVIDA-GTX 1080Ti GPU.

\subsection{Reference Memory}

The CNN provides a bounding box, confidence and class label for each detected object. We discard detections below a confidence threshold (we used a threshold of 0.55 in our experiments). We also discard some object labels we believe are not reliable landmarks such as \textit{person} and \textit{cat}. Each detection is stored as a vector of five floating point values describing the object category and bounding box in the image. We found that each scene might contain two to eight objects, so the resulting whole-scene descriptor is small: of the order of $ 5*8*4 = 160$  bytes. A single original SIFT point descriptor is 512 bytes (both using 32-bit floats). 

\subsection{Relocalization}
\begin{figure}[t!]
	\centering
	\includegraphics[width=0.95\columnwidth]{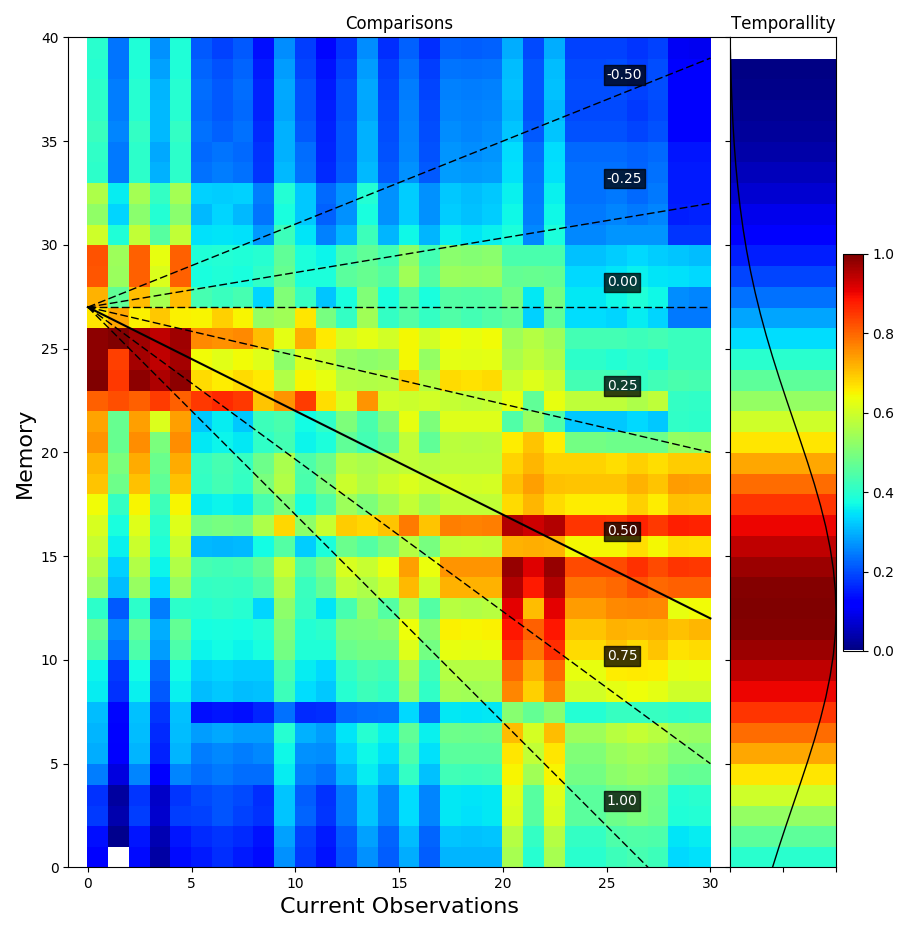}
	\caption{Extended SeqSLAM relocalization method: heatmap of the similarity of a sequence of 10 new scene descriptors compared to memorized descriptors. The line segment with the highest sum of values gives the best estimate of current UAV position and velocity. After initial relocalization we weight nearby scenes to impose a temporality constraint that avoids being confused by time-extended loop-closures. }
	\label{seq_temp}
\end{figure}

After the teaching phase, the reference memory contains a sequence of scene descriptors. Starting from an initial robot pose that can be arbitrarily far from the original trajectory, the robot needs to localize itself in the environment to be able to take a proper action. Given the current observation, it should find the best-matching descriptor in the reference memory.

For matching, we measure the similarity of two individual object detections with matching labels as the ratio of the area of overlap of their bounding boxes to the sum of their  bounding box area:
%% TODO:
\begin{equation}\label{object_matching_score_func}
	S_O = \frac{(\alpha s' \cap \alpha s")}{\alpha (s' + s")}
\end{equation}
Where $ s' $ and $ s" $ denotes the area of the 2D bounding boxes of objects recorded in the target and tested images respectively. $ \alpha $ is a weighting factor to tune matching sensitivity. We chose this by experiment, to get a working value of 4. The similarity of two scenes is computed by averaging the similarity scores of all their objects ($ S_O $): 
\begin{equation}\label{image_matching_score_func}
	S_I = \frac{\sum_{n=1}^{N}{S_O}^{n}}{N}
\end{equation}
where $ N $ is the number of objects in the scene.
% RV: this means area of bounding boxes with matching object categpory, right?

Compared to typical point-feature approaches, we have very few features in each scene, and a dictionary of only a 80 object categories. Thus scene descriptors may not be highly discriminative, e.g. lots of scenes contain a chair below a keyboard below a screen. This problem is also present in methods where low-resolution whole-images are used as decriptors, so we borrow the SeqSLAM technique developed for that domain \cite{milford2012seqslam} where localization is done over a contiguous sequence of images. The last N observations are compared with the memorized observations and a score table is recorded. Since the robot may have different velocities in the teach and repeat phases, we must test different velocities to find a good match. Figure \ref{seq_temp} shows an example score table and line segments corresponding to different UAV velocities. We find the line that passes though score cells with the highest sum. The gradient of the line gives the UAV velocity and the peak value along the line gives the current position along the trajectory.

Testing relocalization, we found that in trajectories with repeated sections (time-extended loop-closures), basic sequence matching could match the wrong location and the robot could become stuck in a loop. To address this, after initial relocalization, we positively weight descriptors nearby in the sequence to add a temporality constraint. With this mechanism, the robot can correctly repeat trajectories with repeated parts. 

Adding the temporality constraint, the similarity measure becomes:
\begin{equation}\label{temporality_eq}
	S_t = \frac{S_s (1 + \beta norm(x=i, \mu=i', \gamma)}{(1 + \beta)}
\end{equation}

Where $ S_s $ is the score of matching that has been done by sequential comparison and $ \beta $ is the factor of temporality. $ i $ is the index of the memory which is being calculated and $ i' $ is the average of the last 5 matched indexes of the memory.

\subsection{Motion Control}

\begin{figure*}[t!]
	\centering
	\includegraphics[width=\textwidth]{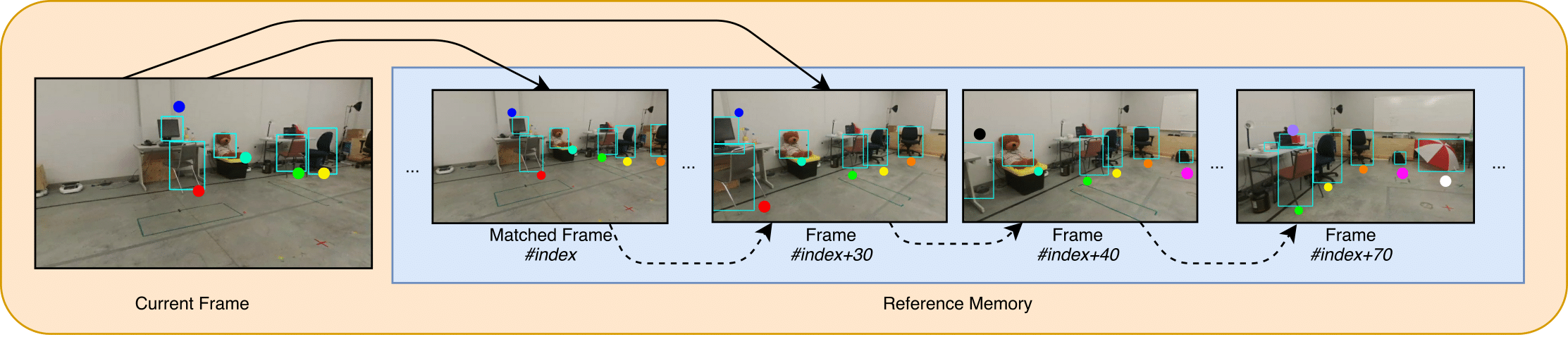}
	\caption{Funnel Lane with short look-ahead window plus Virtual Funnel Lane with long look-ahead window to react to unseen but predicted objects.}
	\label{controller_vo}
\end{figure*}
\begin{figure}[h!]
	\centering
	\includegraphics[width=0.95\columnwidth]{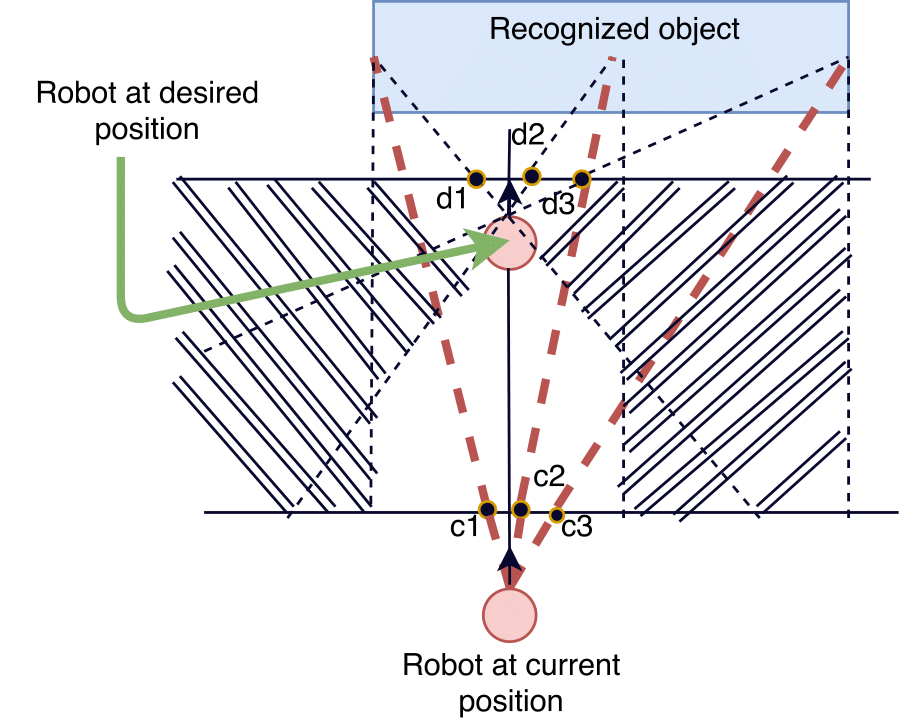}
	\caption{Schematic Funnel Controller}
	\label{controller_schematic}
\end{figure}

Our learned trajectory is in the form of object locations within keyframes, so we use a following controller afforded by this information. `funnel lane' path following controller \cite{chen2009qualitative} method uses image-space point landmarks to achieve robust visual servoing. We extend it here for the case of object landmarks. We control robot pose on a 2D plane of constant altitude only, but the method is trivially extended to altitude control. Three points are are considered for each object detection: one at the center and one at each extreme of its bounding box in the horizontal axis. Using the extent of the object allows us to reason about its scale change (but not its size in 3D). Figure \ref{controller_schematic} sketches the geometry of how two detections of the same object can be used to construct a `funnel lane' that guides the motion of the robot to obtain the second camera position given the first. 

One object detection is enough to repeat an $(x,y)$ trajectory with respect to that object, but the approach direction is not constrained. A second object allows us to control yaw, to get replay of the full sequence of 2D poses $(x,y,\theta)$. 

% RV - is that last part correct?

% RV : TODO
Equation \ref{Funnel_controller} shows the funnel lane controller response for the single landmark on the left x-extent of the recognized object in Figure \ref{controller_schematic}. $ c_1 $ and $ d_1 $ are the horizontal distance of the land mark in image space of the current position and desired position respectively, and $ \phi(c_1, d_1) = \frac{1}{\sqrt[]{2}} (c_1-d_1) $. 

Equation \ref{Funnel_Obj} averages the response for a whole object over its three landmarks. The final response is a command for heading adjustment. We scale the output by a constant gain for tuning (we used a factor of 12). 

\begin{equation}\label{Funnel_controller}
	\theta_{c_{1}} = 
    	\left\{
        	\begin{array}{rcl}
         		 \gamma \min \{c_1, \phi (c_1, d_1)\},    & \mbox{if}  & c_1>0 $ and $ c_1>d_1 \\
                 \gamma \max \{c_1, \phi (c_1, d_1)\},    & \mbox{if}  & c_1<0 $ and $ c_1<d_1 \\
                  0, & &\mbox{otherwise} \\
            \end{array}
        \right.
\end{equation}

\begin{equation}\label{Funnel_Obj}
\theta_{obj}=\frac{(\theta_{c_1}+\theta_{c_2}+\theta_{c_3})}{3}
\end{equation}

% RV TODO
Often, several objects are detected in a single scene. When this occurs, the controller calculates the response for each object individually using Equation \ref{object_matching_score_func}. Then it performs the funnel lane method for all objects as sketched in \ref{Funnel_Obj} (a full desription is ommitted for space).

\begin{equation}\label{Funnel_Image}
	\theta_{image}=\frac{\sum_{n=1}^{N}{\theta_{obj}}}{N}
\end{equation}

% please try to improve the language here: I don't understand it
%\subsubsection{Looking ahead in the learned trajectory}
%The controller fetches the next ten sequential key-frames from the memory starting from the currently-matched frame. It compares those frames in the memory with the current observation and makes a linear weighted average which emphasizes the more distant key-frames. Thus the robot is able to look ahead, making it more agile specifically in sharp turns. % Although the number of future frames could be varied by different velocities at the teaching phase, it could either be fast and objects in the first frame go out of the image after a few frames or too slow which makes memory have redundant observations. We found out that ten future frames are enough to take reasonable actions, cause having a large number may leads the controller to take the wrong action.

%Also, to have more information about the future of the trajectory, visual odometry based on the teaching route can be a good alternative. Since, in this paper, we are persisting to use semantic information, 

The basic funnel controller has a failure mode that is difficult for our application: it steers the robot to align the currently-perceived objects as they should appear in later frames. But it does not react to upcoming objects that can not yet be seen. Thus with a narrow field of view camera it fails to turn the robot in sharp corners. To address this, we augment the funnel lane controller with what we will call a \textit{virtual funnel lane}: we simulate the action of the normal funnel lane controller by running it on the sequence of stored images from the current best match up to some window lookahead size. The robot simulates what will happen in the future, assuming it is correctly localized. This was the robot can `see around' upcoming corners. The robot records the heading changes prescribed by the controller in this virtual look-ahead flight, and averages them to obtain a predicted future yaw value. This is then blended with the original funnel lane control by Equation \ref{Funnel_total}. The intuitive effect of these two control components is that the `real' funnel lane aligns the robot well to the things it can see now, while the virtual funnel lane pre-aligns the robot to objects it can not yet see. Thus the robot will turn nicely around corners even when the set of objects in view changes completely. A detailed analysis of this method is beyond the scope of this paper.

\begin{equation}\label{Funnel_odometry}
	\theta_{Virtual}= \frac{\sum_{i=I_m}^{W_v}{\theta_{image} (i,i+1)}}{W_v}
\end{equation}

\begin{equation}\label{Funnel_Funnel}
	\theta_{Funnel}= \frac{\sum_{i=I_m}^{W_f}{\theta_{image} (current,i)}}{W_f}
\end{equation}

where $ I_m $ is the index of the matched keyframe and $ W_v $ and $ W_f $ denote the size of the virtual funnel window and funnel control window respectively. We used $ W_v = 70 $ and $ W_f = 30 $. 

\begin{equation}\label{Funnel_total}
	\theta_{total}=\alpha \theta_{funnel} + (1-\alpha)\theta_{virtual}
\end{equation}
Where $ \theta_{funnel} $ 
This enhanced controller is illustrated in Figure \ref{controller_vo}. The current image is only being compared with the next 30 key-frames in the memory to generate a funnel lane control command, but the virtual funnel lane looks ahead 70 key-frames to consider objects not yet in view such as the umbrella. The look-ahead  anticipates the turn towards the umbrella and forces the controller to change the robot's heading towards it.

To complete control logic is: (i) rotate on the spot if not well localized; (ii) if yaw error is below a threshold, go forward with constant velocity, else yaw to correct the error. We decoupled yaw and forward motion as the low-level flight controller behaved badly when attempting to yaw and translate at the same time due to the complex vehicle and controller dynamics of a quadrotor.

%There is another term in Equation .. which is relevant to odometry and can be added to control command by weigthed average. 
%VO=...
%control command= ..
%average N, N_VO number of odometery window.

%% Second Version

\section{Experiments}
% TODO: experiment setup should be described here

\begin{figure}[t!]
	\centering
	\includegraphics[width=0.95\columnwidth]{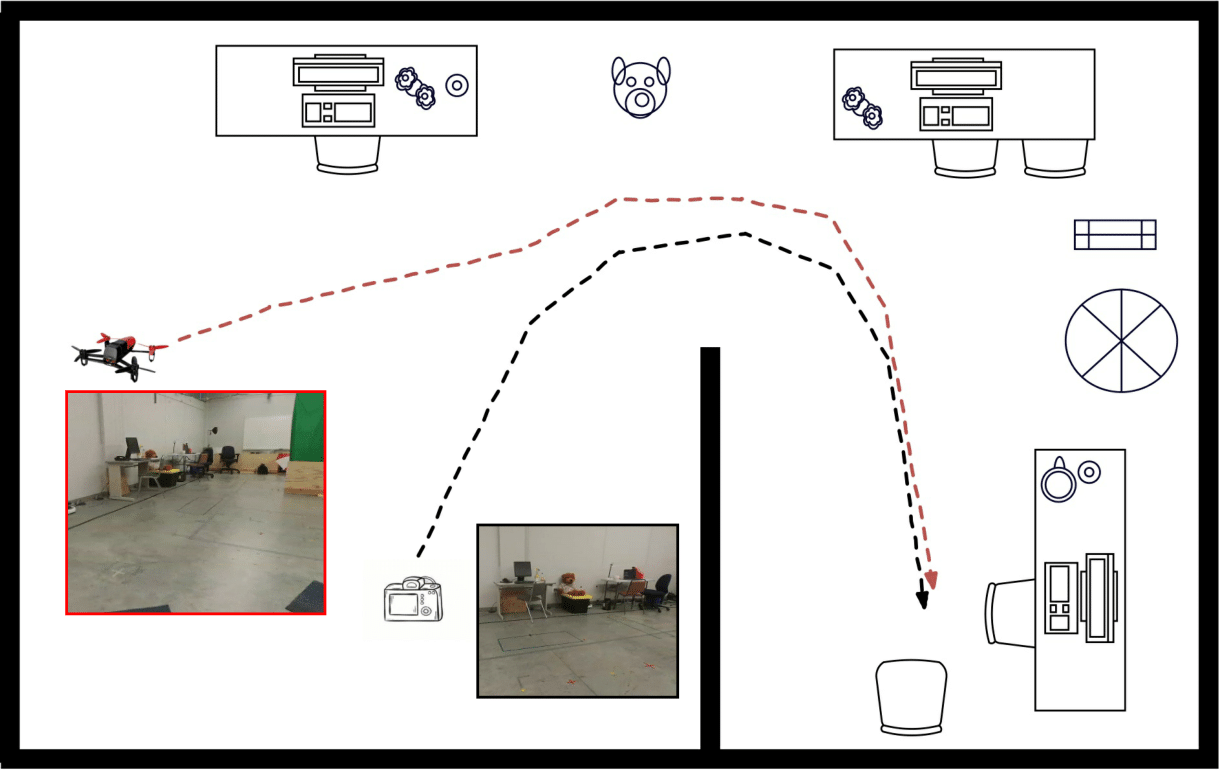}
	\caption{Test scenario: The trajectory of the camera is encoded as a series of object landmark locations in image space. Later, the UAV autonomously localizes itself in this space, then repeats the remaining part of the camera trajectory. A CNN detects objects such as desk, chair, mug, umbrella, backpack in keyframes. The UAV trajectory is generated directly from pairs of keyframes.} 
	\label{schematic}
\end{figure}

We performed real-world experiments with a commodity Bebop 2 UAV. We are limited to a 10m x 6m experimental arena where we can independently measure UAV trajectories using a Vicon motion-capture system. The test environment is pictured in schematic in Figure \ref{schematic}. We limited the total number of recognizable objects in the environment to 18.

\subsection{Training}
In the training phase, the UAV is used as a passive camera. It is placed on a wheeled cart and pushed along the desired route, indicated by the trajectory marked with a camera Figure \ref{schematic}. The altitude is constant at 1.5m. The pose is recorded independently by motion capture. The video is fed to the CNN and descriptor generator and the sequence of memory of 290 key-frames is built. 
%Picture of cart and training data procedure

\begin{figure}[t!]
	\centering
	\includegraphics[width=\columnwidth]{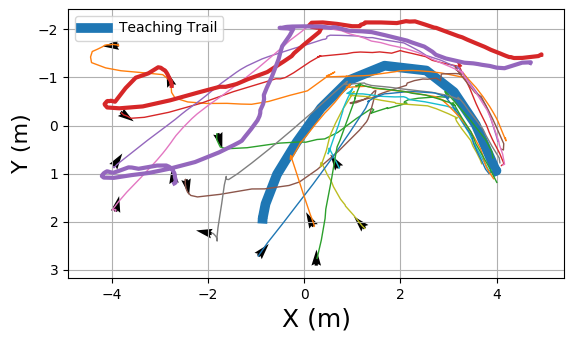}
	\caption{Trails in both teaching and repeating phases, arrows are indicating initial rotation of the robot.}
	\label{Experiment2_result}
\end{figure}

\subsection{Experiment I: Changing Viewpoint}
A first experiment was performed to test the robustness of relocalization and trajectory repeating to changes in initial viewpoint. We run 12 trials with the the robot starting from $(x,y,\theta)$ poses chosen at random up to 5m from the original camera, and one of two fixed altitudes chosen at random. The robot flies autonomously, controlled by the VTR system. On startup, the robot rotates on spot until relocalization is achieved, then attempts to repeat the trajectory until matching the last keyframe.

\subsubsection{Results}
Out of 12 trials, 10 robots completed the learned trajectory, arriving less than 0.5m from the original end-point. Figure \ref{Experiment2_result} reports the start positions and the paths of all robots. The blue bold line is the taught path. Note that the robot navigates around an opaque wall (Fig \ref{schematic}), so that the features detected in the second half of the trajectory are not visible in the first half: direct visual servo to the final target can not solve this task. In both the failed cases, the robots incorrectly detected the endpoint keyframe too early, stopping at the pair of chairs at the top right, and not the correct final pair of chairs at the bottom right. Both did correctly localize initially and completed 65\% of the trial correctly.

%picture from the lab with blue carpets showing the spots.

\subsection{Experiment II: Changing Environment}
A second experiment investigates the ability to repeat a learned trajectory when the environment changes. The robot starts approximately at the same place as in teaching. But in one set of trials we change the lighting by turning off the ceiling lights and turn on two spot lamps. In a second set of trials we remove five objects that were noted by the object detector in training, and move eight others. The supplementary video shows us disturbing the objects to change their appearance.  We repeat the trajectory and record the time taken to arrive at the final landmark. 

\begin{figure}[t!]
	\centering
	\includegraphics[width=0.95\columnwidth]{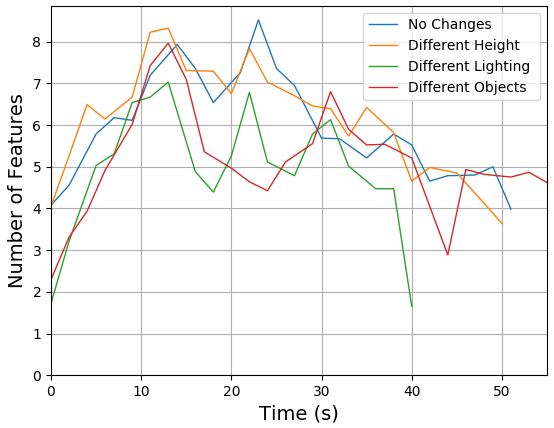}
	\caption{Number of features detected at nearest-corresponding locations during teaching and repeat phases.}
	\label{features_counter}
\end{figure}

\subsubsection{Results}
\begin{figure}[t!]
	\centering
	\includegraphics[width=0.95\columnwidth]{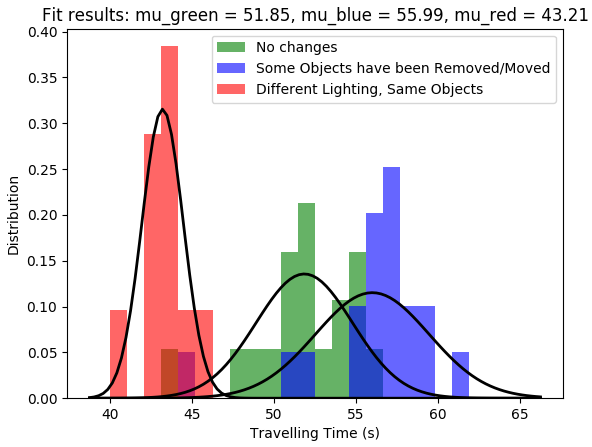}
% RV TODO: fix colors in next sentence
\caption{Repeat completion times for identical environment (color), light-changed environment (color) and object-moved environment (color).}
	\label{Normality_Figure}
\end{figure}

Examining robustness to appearance changes, we record the time taken to repeat the trajectory after (a) drastically changing the lighting and (b) removing objects and moving others to change their appearance. A histogram of flight times is shown in figure \ref{Normality_Figure} with fitted Gaussians. 

Removing and moving objects changes the execution time distribution, increasing it by an mean of 4sec (less than 10\%).

But, perhaps surprisingly, changing the illumination reduces time taken for a repeat. The data show that the changed illumination slightly reduced the number of recognized objects, in particular the objects that were less repeatably detected in the teaching phase. Without objects disappearing and reappearing as frequently, the flight controller produced smoother behaviour and reached the goal faster. Perhaps the method could be improved by deliberately filtering such 'flickering' objects. 

Figure \ref{features_counter} provides evidence of this effect, showing that the number of detected objects differs during the light-changing and removing/moving objects experiments. When removing/moving objects, the number of detected features decreases, but the run time is longer because the moved objects degrade the controller behaviour, and the robot takes longer to line itself up along the trajectory. In the spot-lighting trials, the fewer object detections improve completion time as described above.

% RV TODO missing ref []
\section{Conclusion, limitations and future work}
In this paper we showed that a current deep-CNN object detector can be used to provide robust and repeatable features that are sufficient for monocular VTR. We demonstrated the method in real-world UAV experiments. Objects are detected with viewpoint and lighting invariance that compares favourably with other methods, and with interesting novel invariances to the object being completely turned around or replaced by another of the same category. We demonstrated that we can start the robot 5m away from the taught initial position, facing at a random heading, and the robot could relocalize itself and complete the trajectory using the object detections alone. Comparable experiments using SURF features have been shown to work at not more than 1m disturbance. We also successufully repeated a trajectory with the UAV's altitide changed by 40\%, and also if several objects are removed and removed, or if the global lighting is changed dramatically. 

We also contribute a novel two-window use of the funnel lane visual servo method that looks ahead in time to respond to upcoming objects. 

Our method depends on a supply of distinct ambient objects. It will not work in environments without discontinuous, recognizable object categories. Previous work on automatic generation of low-level feature dictionaries could suggest directions for future work. We aim to take this work outdoors over kilometer scales. For this, we need a CNN that reliably detects objects that exist in our target environment. We may have to train our own networks on the salient objects.

% We did another experiment to show how algorithm is robust to illuminational changes and object removal or moving in the scene. It is something that might happen regularly in every environment and robot should still be able to repeat the taught path. By normal distribution test, we showed that algorithm is robust enough to endure .. object removal/moving and harsh lighting changes. mean value of completion time is almost near the original environment test (52s for original, 56s for object removal/move and 46s for illuminational change).

%In future of this work, one can use other information in the scene, not only objects, to make this algorithm more robust for relocalization, even in the environments without rich object. In this order, using deep network representation is an advantage and segmenting networks can be a beneficial tools, specially for outdoor environment. Asking the robot to do a specific task like take a close-up photo from a specific object is another potential extension of this project.  

One simple but interesting extension of the work would be for the robot to announce changes to the set of objects. If a previously-seen vase or painting is missing, perhaps it has been stolen. If a teddy bear has appeared, perhaps its owner wants it back. More substantially, perhaps a robot is performing an inspection trajectory over a new manufactured object such as an aeroplane wing. Is a rivet missing or has a bad weld appeared, compared to the teach phase recorded over a known-good example?

The recent performance improvement in some computer vision tasks due to CNNs and similar methods are giving us interesting new choices for robot vision. We aim to exploit the new  robust vision methods towards more complete robot autonomy and new applications.


\section*{ACKNOWLEDGMENT}
Supported by the NSERC Canadian Field Robotics Network.

%%%%%%%%%%%%%%%%%%%%%%%%%%%%%%%%%%%%%%%%%%%%%%%%%%%%%%%%%%%%%%%%%%%%%%%%%%%%%%%%

% USE BIBTEX rather than formatting your own refs. You won't get it right.
% refs take a LONG TIME. 

\bibliographystyle{IEEEtran}
\bibliography{ref}

\end{document}